%% file: main.tex
\documentclass[11pt]{article}

\usepackage[preprint]{acl}

\usepackage{times}
\usepackage{latexsym}

\usepackage[T1]{fontenc}

\usepackage[utf8]{inputenc}

\usepackage{microtype}

\usepackage{inconsolata}

\usepackage{graphicx}

\usepackage{booktabs}
\usepackage{multirow}

 \usepackage{amsmath}

\usepackage{tikz}
\usetikzlibrary{shapes.geometric, arrows.meta, positioning, fit, backgrounds, calc}

\usepackage{listings}
\input{custom_commands}

\title{MASEval: Extending Multi-Agent Evaluation from Models to Systems}

\author{
  \textbf{Cornelius Emde}\textsuperscript{1,2}\thanks{\texttt{ai@corneliusemde.com}},
  \textbf{Alexander Rubinstein}\textsuperscript{3,\S},
  \textbf{Anmol Goel}\textsuperscript{1,4,\S},
  \textbf{Ahmed Heakl}\textsuperscript{1,5,\S},\\
  \textbf{Sangdoo Yun}\textsuperscript{6},
  \textbf{Seong Joon Oh}\textsuperscript{1,3,7}\thanks{\texttt{coallaoh@gmail.com}},
  \textbf{Martin Gubri}\textsuperscript{1}\thanks{\texttt{martin.gubri@parameterlab.de}}
  \\
  \S{}Equal contribution.
  \textsuperscript{1}Parameter Lab,
  \textsuperscript{2}University of Oxford,
  \textsuperscript{3}University of Tübingen, \\
  \textsuperscript{4}TU Darmstadt,
  \textsuperscript{5}MBZUAI,
  \textsuperscript{6}NAVER AI Lab,
  \textsuperscript{7}KAIST
}

\begin{document}

\maketitle

\begin{abstract}
The rapid adoption of LLM-based agentic systems has produced a rich ecosystem of frameworks (smolagents, LangGraph, AutoGen, CAMEL, LlamaIndex, i.a.). Yet existing benchmarks are model-centric: they fix the agentic setup and do not compare other system components. We argue that implementation decisions substantially impact performance, including choices such as topology, orchestration logic, and error handling. MASEval addresses this evaluation gap with a framework-agnostic library that treats the entire system as the unit of analysis.
Through the first systematic system-level comparison across 3 benchmarks, 3 models, and 3 frameworks, we find that, across models within a capability tier, framework choice matters as much as model choice.
MASEval allows researchers to explore all components of agentic systems, opening new avenues for principled system design, and practitioners to identify the best implementation for their use case.
MASEval is available under the MIT licence at \href{https://github.com/parameterlab/MASEval}{github.com/parameterlab/MASEval}.
\end{abstract}

\section{Introduction}

LLM-based agentic systems have become popular for automating complex workflows. Single-agent architectures use one LLM equipped with tools to iteratively reason toward a solution. Yet as workflows grow in complexity, the field is increasingly adopting multi-agent systems, where specialised agents collaborate through structured coordination. This shift promises improved task decomposition and cost efficiency through smaller, specialised models \citep{belcak2025small}. The growing adoption of multi-agent approaches is reflected in the proliferation of frameworks such as AutoGen \citep{wu2024autogen}, LangGraph \citep{langgraph}, and CAMEL \citep{li2023camel}, which embody different architectural choices for agent communication, memory management, and coordination.

\input{assets/figures/motivating_figure}

This transition from single-agent to multi-agent systems fundamentally changes evaluation requirements. For single-agent setups, benchmarks could reasonably focus on model capabilities within a fixed agentic scaffold. Multi-agent systems, however, introduce new dimensions that demand attention, including organisational topologies, communication protocols, memory architectures, role differentiation, and the orchestration logic that governs their interaction \citep{guo2024survey}. Moreover, the proliferation of frameworks raises a question that current benchmarks cannot answer: which framework should I use?

\input{assets/tables/library_comparison}

Existing benchmarks such as GAIA \citep{mialon2024gaia} and AgentBench \citep{liu2023agentbench} remain predominantly model-oriented and static. The typical benchmark report states `GPT-4 achieves 85\% on task X,' which conflates model capability with framework implementation. Evaluation libraries for single-agent setups also lack infrastructure to trace multi-agent coordination. This leaves fundamental questions unanswered: is multi-agent topology better than single-agent for task X? Which tool-calling format? Does smolagents outperform LangGraph?

This evaluation gap has concrete consequences. \emph{Researchers} lack a principled way to compare design decisions such as communication topologies or coordination strategies and cannot easily build on prior architectural findings. \emph{Practitioners} receive no data-driven guidance on framework choice. \emph{Benchmark consumers} face fragmented interfaces, as evaluating a system across multiple benchmarks requires significant boilerplate reimplementation. \emph{Benchmark producers} must reinvent increasingly complex evaluation infrastructure for each new benchmark.

We introduce MASEval, the first \textbf{framework-agnostic evaluation library for multi-agent systems} (Figure~\ref{fig:motivation}). Our key principle is to evaluate the complete system (agents, framework, and coordination logic). MASEval provides:

\begin{enumerate}
    \item \textbf{System-level evaluation infrastructure} for comparing design decisions and framework implementations. Both matter for building effective multi-agent systems.
    \item \textbf{Unified benchmark interface} for evaluating agents across multiple benchmarks with minimal integration overhead.
    \item \textbf{Benchmark development toolkit} for creating new benchmarks without reinventing evaluation boilerplate.
    \item \textbf{Multi-agent tracing} with per-agent message histories for debugging coordination patterns.
\end{enumerate}

To validate MASEval's utility, we conduct experiments across 3 benchmarks, 3 frameworks, and 3 models. Our key finding: \textbf{framework choice impacts performance comparably to model choice}. This result was previously obscured by existing model-centric benchmarks, and demonstrates the importance of system-level evaluation in MASEval's design. MASEval reduces implementation effort by 83--91\% for benchmark consumers who adopt existing benchmarks and by 35--57\% for benchmark producers who build new evaluations.


\section{Related Work}
\label{sec:related-work}

MASEval is at the intersection of agent frameworks that build systems, benchmarks that define tasks, and evaluation libraries that measure performance.

\noindent \textbf{Multi-agent frameworks.} LLM-based agent frameworks span academic systems \citep{li2023camel,hong2024metagpt,chen2024agentverse}, developer-oriented tools \citep{wu2024autogen,langgraph,llamaindex2022,crewai,smolagents,pydantic_ai}, and frontier-provider SDKs \citep{openai_agents_sdk,google_adk,claude_agent_sdk}. Frameworks differ in design philosophy: stateless vs.\ stateful execution, static vs.\ dynamic control flow, centralised vs.\ decentralised communication, and JSON-based vs.\ code-based tool calling. AnyAgent \citep{any_agent} unifies execution across frameworks but does not address evaluation. No prior work provides infrastructure for cross-framework \emph{evaluation} that decouples the system under test from the benchmark harness.

\noindent \textbf{Evaluation libraries.} Inspect AI \citep{inspectai2024} offers a convenient, batteries-included evaluation framework with pre-built benchmarks, but its solver abstraction has no notion of multiple cooperating agents, making per-agent tracing with arbitrary agent implementations non-trivial. HAL \citep{kapoor2026hal} is limited to single-agent and locks logging to W\&B Weave. MARBLE \citep{zhu2025multiagentbench} supports multi-agent coordination but restricts topologies to a predefined set and requires agents to be implemented within its own paradigm, precluding the use of custom frameworks. LLM observability tools \citep{phoenix2022,trulens2020,zaharia2018mlflow} focus on monitoring rather than benchmark execution. Commercial platforms \citep{langsmith2024,braintrustsdk2023,galileopython2025} introduce vendor lock-in. None supports evaluating multi-agent systems across frameworks with per-agent tracing. Table~\ref{tab:comparison} compares MASEval with other libraries.

\noindent \textbf{Benchmark datasets.} Agent benchmarks cover single-agent capabilities \citep{jimenez2024swebench,mialon2024gaia,barres2025tau2,liu2023agentbench}, multi-agent collaboration \citep{zhu2025multiagentbench,shu2024macs,xu2024theagentcompany,froger2025are}, and safety \citep{vijayvargiya2026openagentsafety,gomaa2025converse}. Each ships with an incompatible evaluation interface, requiring bespoke integration code. MASEval's unified benchmark interface addresses this fragmentation.

\section{System Architecture}
\label{sec:architecture}

MASEval bridges the gap between frameworks and benchmarks. It is neither an agent framework nor a benchmark dataset, but an \emph{evaluation infrastructure} that enables any agent to be evaluated on any benchmark through a universal interface. 

\input{assets/figures/byo_philosophy}

\subsection{Design Principles}
\label{sec:principles}

The challenges from \S\ref{sec:related-work} motivate five design principles that distinguish MASEval from prior evaluation approaches:

\begin{enumerate}
    \item \textbf{System as unit of analysis.} The agent system is evaluated as a whole, enabling comparison of architectural choices and framework implementations, not just model capabilities.

    \item \textbf{Bring your own.} No framework, model provider, or logging backend is privileged. The core never imports framework-specific code (enforced via CI). Pre-built adapters exist for common choices.

    \item \textbf{Infrastructure, not implementation.} MASEval provides orchestration, tracing, and lifecycle management; users control tools, agent behaviour, and evaluation metrics. Like PyTorch Lightning, it reduces boilerplate without abstracting away evaluation logic.

    \item \textbf{Separation of concerns.} Task definition (what to solve), environment (tools and state), agent logic (how to solve), and evaluation (how to measure) are cleanly separated. Each component can be varied independently to isolate its effect on performance.


    \item \textbf{Trace-first evaluation.} All components log to a shared trace organised by component. Agent messages, model usage, and tool calls can be directly inspected. Per-agent traces are kept independent to support partial observability.
\end{enumerate}

These principles involve deliberate tradeoffs. Agent systems are architecturally diverse. One is a Python library call, another a command-line tool, another a containerised service or remote endpoint. Any assumption about agent internals therefore excludes valid implementations. The ``Bring Your Own'' philosophy keeps the adapter contract minimal (run the agent and retrieve its messages) because that is all one can assume. The cost is higher initial setup compared to opinionated libraries. The gain is flexibility: wrapping a containerised service is as straightforward as wrapping an in-process framework agent.

This flexibility extends to infrastructure. Existing evaluation codebases couple the original authors' choices (a specific cloud platform, a hard-coded provider SDK) throughout the code. A team without access to that platform must rewrite far beyond the evaluation logic. MASEval isolates these choices behind pluggable interfaces. The cost is a thinner per-platform feature surface, forgoing turnkey dashboards or native experiment tracking. The gain is that no team is locked out by a dependency they cannot use, and swapping a backend is a single-line change.

Finally, we keep the abstraction surface minimal. In production systems, robustness means defensive coding with deep exception hierarchies and exhaustive validation. In a research context, such patterns obscure the evaluation logic and mask rather than surface problems. For MASEval, robustness means reliable, reproducible results with code that remains readable and easy to extend.

\subsection{Module Architecture}
\label{sec:modules}

MASEval realises these principles through a module structure with strict dependency boundaries that keep evaluation logic independent of any particular framework, model provider, or logging solution (Figure~\ref{fig:byo-philosophy}).

\noindent \textbf{Core \code{maseval/core/}.} Abstract base classes and evaluation runtime defining the contracts (\code{AgentAdapter}, \code{Environment}, \code{Evaluator}, \code{ModelAdapter}, \code{User}). Orchestrates the benchmark lifecycle with minimal dependencies.

\noindent \textbf{Interface \code{maseval/interface/}.} Lazy-loaded adapters for frameworks (smolagents, LangGraph, LlamaIndex), model providers (OpenAI, Anthropic, Google), and logging destinations. Supporting a new framework requires implementing only a thin adapter.

\noindent \textbf{Benchmark \code{maseval/benchmark/}.} Complete benchmark implementations (e.g., $\tau^2$-Bench, MultiAgentBench) serving as both ready-to-use evaluations and reference implementations.

\subsection{Core Abstractions}
\label{sec:abstractions}

MASEval defines seven core abstractions that benchmark producers implement and benchmark consumers rely on:

\noindent \textbf{Task.} The atomic unit of evaluation: bundles the query, environment data, evaluation criteria, and execution protocol.

\noindent \textbf{Benchmark.} Orchestrates the evaluation lifecycle for a task collection. Users subclass \code{Benchmark} and override hooks (\code{setup\_environment}, \code{setup\_agents}, \code{setup\_evaluators}, etc.) while inheriting execution and tracing infrastructure.

\noindent \textbf{Environment.} Manages state and exposes tools that agents can invoke, remaining stateful across turns within a task execution.

\noindent \textbf{AgentAdapter.} Wraps any framework's agent in a standard interface. Exposes message history for tracing.

\noindent \textbf{User.} Simulates user responses for multi-turn benchmarks, with configurable personas and turn limits.

\noindent \textbf{Evaluator.} Computes metrics via a two-stage pattern: filter traces to extract relevant data, then compute metrics.

\noindent \textbf{ModelAdapter.} Unified interface for LLM providers with tool calling and token tracking. Used by simulators and LLM-based evaluators; agents use their framework's native integration.

\subsection{Benchmark Lifecycle}
\label{sec:lifecycle}

\input{assets/figures/benchmark_lifecycle}

Each task execution proceeds through five phases (Figure~\ref{fig:benchmark-lifecycle}), iterated over tasks and repetitions (\code{n\_task\_repeats}):

\begin{enumerate}
    \item \textbf{Setup.} Instantiate environment, tools, user simulator, agents, and evaluators; register all components for automatic trace collection.
    \item \textbf{Execute.} Run agents with customisable turn orchestration.
    \item \textbf{Collect.} Gather traces from all registered components and system metadata.
    \item \textbf{Evaluate.} Compute metrics from full traces, including intermediate steps and tool usage.
    \item \textbf{Report.} Store a structured report (task ID, status, traces, configuration, results) and clear the registry for the next repetition.
\end{enumerate}

\noindent Implementing multiple benchmarks with diverse designs (Table~\ref{tab:supported_benchmarks}) has validated that this phase ordering and separation of concerns hold across different benchmark designs.

\subsection{Key Capabilities}
\label{sec:capabilities}

\noindent \textbf{Multi-agent tracing.} Each agent maintains an independent message history; the component registry automatically collects per-agent traces, enabling debugging of coordination failures.


\noindent \textbf{Callback system.} Lifecycle hooks at benchmark, environment, and agent levels enable extensible monitoring without modifying core logic. Built-in callbacks cover progress bars and result logging. Users can add custom callbacks for experiment tracking, early stopping, or external platform integration.

\noindent \textbf{Adaptive testing.} 
The cost of evaluating modern LLMs can be dramatically reduced by selecting only the most informative benchmark tasks rather than evaluating on the entire dataset~\citep{vivek-etal-2024-anchor,tinybenchmarks, rubinstein2026disco}. The \code{AdaptiveTaskQueue} enables this by selecting the next task adaptively based on the model's current estimated skill, for instance using Item Response Theory~\citep{lord1952theory,rasch1960probabilistic, lord1968statistical}. The \code{InformativeSubsetQueue} interface supports various strategies to pre-select subsets of tasks from which overall performance can be estimated. For example, the DISCO algorithm~\citep{rubinstein2026disco} can estimate overall benchmark performance within roughly $2\text{pp}$ of a full evaluation while using only $1\%$ of all available tasks. This compute reduction is particularly valuable for frontier models, where a single benchmark run can cost tens of thousands of dollars\footnote{\url{https://x.com/DimitrisPapail/status/2026699305021587641}}.


\noindent \textbf{Reproducibility infrastructure.} Every report includes git state, system information, and package versions; \code{ConfigurableMixin} extends this to any user-defined component.


 

\subsection{Key Features}

MASEval provides production-quality infrastructure for evaluation: parallel task execution, structured error attribution, pluggable logging backends, adaptive task scheduling, and reproducibility tooling. Appendix~\ref{sec:features} lists key features. \mbox{MASEval} is installable via \code{pip install maseval} (MIT licence). The documentation at \href{https://maseval.readthedocs.io}{maseval.readthedocs.io} covers API reference, implementation examples, and design patterns.

\section{System-Level Benchmarking} 
\label{sec:experiments}

Current benchmarks focus on model comparison, neglecting \emph{framework implementations} and \emph{design decisions}. We use MASEval to conduct a systematic cross-framework comparison that holds the agent architecture constant and varies only the framework and model. Consider a practitioner who has a multi-agent design in mind and must choose both a model and a framework to implement it. Our experiment measures exactly this: given the same design implemented idiomatically in three frameworks and powered by three models, what performance differences emerge out of the box? We find that framework-level design choices can match or exceed the performance differences between models of comparable capability.

\subsection{Experimental Setup}

We conduct a full factorial experiment across 3 frameworks, 3 models, and 3 benchmarks (27 configurations). For each benchmark we select 2 domains and run all tasks within each domain.

\input{assets/tables/framework_performance.tex}

\noindent \textbf{Benchmarks.} We select three benchmarks spanning capability and safety evaluation in multi-agent settings.
\textbf{MACS} \citep{shu2024macs} tests multi-agent coordination on enterprise tasks. We report partial goal success rate (pGSR).
\textbf{\textsc{ConVerse}} \citep{gomaa2025converse} measures resistance to security attacks in agent-to-agent conversations. We report robustness ($1-\text{ASR}$). The attacker agent is held constant and only the defending agent varies across frameworks.
\textbf{MultiAgentBench} \citep{zhu2025multiagentbench} evaluates both collaboration and competition between agents. We report Task Completion Rate for the Research domain and scaled Task Scores (TS) for Bargaining.

\noindent \textbf{Frameworks.} We compare three frameworks through MASEval's built-in adapters. For \textbf{smolagents} \citep{smolagents} (minimalist, code-based tool calling) we wrap \code{ToolCallingAgent} via \code{SmolAgentAdapter}; for \textbf{LangGraph} \citep{langgraph} (state machine-based coordination with explicit state management) we wrap \code{StateGraph} via \code{LangGraphAgentAdapter}; and for \textbf{LlamaIndex} \citep{llamaindex2022} (flexible single- and multi-agent framework) we connect \code{FunctionAgent} with \code{AgentWorkflow} via \code{LlamaIndexAgentAdapter}.

\noindent \textbf{Models.} We test three current-generation, cost-efficient mid-tier models from different providers: GPT-5-mini, Gemini-3.0-Flash, and Claude-Haiku-4.5. All sit at a similar price and latency point, representing a realistic set of alternatives a practitioner faces alongside the choice of framework. Comparing across tiers (e.g.\ GPT-5-mini against GPT-5-Pro) would likely yield larger model effects but also shift focus towards the cost-performance frontier, complicating the analysis. All models are used with temperature $1.0$ and top-p $1.0$ as these are universally available across all models.

\noindent \textbf{Controlled variables.} We aim to isolate the design choices made by each framework while keeping all other variables constant. We run each benchmark with hyperparameters chosen to replicate the original paper and applied uniformly across conditions, including tool definitions, agent topologies, user simulations, environment dynamics, evaluation logic, and execution limits (e.g., maximum user turns, step budgets). Where frameworks differ in step granularity (e.g., one step accounting for tool call and response vs. two separate steps), we map limits accordingly. For judges, environment simulation, and attackers, we utilise the Gemini-3.0-Flash model across all conditions. Only the framework varies, together with its built-in defaults: system prompts, tool-mounting mechanisms, and error handling. We do not modify or align these internals across frameworks.

\subsection{Results}

Agentic evaluations are inherently noisy due to stochastic model outputs and compounding decision paths, so we focus on aggregate patterns rather than definitive pairwise rankings. Table~\ref{tab:results} presents our main results. To compare the relative impact of framework and model choice, we quantify the variability either choice induces. The bottom four rows capture how much performance shifts when each factor is swapped while holding the other fixed, reporting both range and standard deviation of that spread. Each value averages over the complementary axis: the cross-model range in a given domain, for instance, is the mean of the best-minus-worst model score taken within each of the three frameworks.

\noindent \textbf{Framework impact is substantial.} Framework choice produces performance differences comparable in magnitude to model choice within the same capability tier. Across the six domains, the range between frameworks exceeds that between models in two of the six domains. Averaging over all six domains, the mean range is 14.2 percentage points (pp) across models and 12.4~pp across frameworks, with mean standard deviations of 7.5~pp and 6.5~pp respectively. The most striking single-cell example is Haiku~4.5 on MACS Travel, which scores 90.4 with smolagents but 59.5 with LlamaIndex, a 30.9pp gap between frameworks. Practitioners who tune only the model are therefore optimising part of the system.

\noindent \textbf{Framework-model interactions.} No single framework dominates across all models. On MACS, smolagents achieves the highest scores with Haiku~4.5 but the lowest with GPT-5-mini. To investigate such a strong difference, we carefully analysed traces for smolagents running on MACS and established surprising results. GPT-5-mini struggles with smolagents as the framework forces a tool call at every step, causing GPT-5-mini to overengage the user through repeated clarification attempts. On MACS, where the number of clarifying questions is capped at five, GPT-5-mini misinterprets the ``max turns reached'' error message and retries the same tool call up to 23 times, only rephrasing the question rather than adjusting its strategy in response to the systemic failures of this tool call. While it can finally achieve its goals, it often does so with $\geq10 \times$ higher token consumption compared to other models. Other models and frameworks do not exhibit this failure mode. These interactions illustrate how framework conventions (mandatory tool calling, error message format) can combine with model tendencies to produce failures that neither component would exhibit in isolation.

\noindent \textbf{Implications.} These results validate MASEval's core premise that framework choice matters for agent performance. Model-only evaluation serves model developers, but practitioners and researchers building agentic systems need more: they must also evaluate the orchestration harness and its interaction with the model. \textit{System-level evaluation infrastructure like MASEval is necessary for the field to make informed architectural decisions}.

\subsection{Implementation Effort}
\label{sec:effort}

To quantify MASEval's value as benchmark infrastructure, Table~\ref{tab:loc} compares the lines of code required to define and run benchmarks in MASEval against their original codebases. We report evaluation-related logic, excluding data and prompt templates that remain constant across implementations, adjusted for formatting, docstrings, and ablation code. We focus on \textsc{ConVerse} and $\tau^2$-bench as complete reimplementations in MASEval that do not rely on vendored dependencies.

The most consistent reduction appears in the orchestration layer. The \emph{interface} code that wires components into a runnable benchmark shrinks by 83--91\%, as MASEval provides lifecycle management, trace collection, and execution infrastructure that original implementations build from scratch. This is the key gain for benchmark \emph{consumers}, who need only the interface to run existing benchmarks. Benchmark \emph{definitions} vary depending on the original codebase's structure: $\tau^2$-bench sees a 49\% reduction because its original implementation embedded orchestration logic within component definitions, whereas \textsc{ConVerse}'s cleanly factored design yields nearly identical definition sizes. Overall, total implementation effort decreases by 35--57\%, allowing benchmark producers to focus on evaluation logic rather than orchestration boilerplate.

\input{assets/tables/loc_comparison}

\vspace{-3pt}
\section{Conclusion}
\label{sec:conclusion}
\vspace{-3pt}

MASEval provides framework-agnostic evaluation infrastructure that treats the complete agent system as the unit of analysis. Through multi-agent tracing, structured error attribution, and a unified benchmark interface, it enables systematic comparison of architectural choices and framework implementations. Our experiments show that, within a capability tier, framework choice impacts performance comparably to model choice, challenging the model-centric status quo. System-level evaluation is essential for moving beyond ad-hoc implementations towards principled system design. We welcome community contributions of additional framework adapters and benchmark integrations.


\section*{Ethics Statement}

MASEval is an evaluation infrastructure that does not introduce new capabilities for language models. The benchmarks we implement were designed by their original authors with appropriate ethical considerations. By lowering the barrier to systematic benchmarking, MASEval could accelerate the development of more capable autonomous agent systems. We believe this risk is outweighed by the benefit of enabling the safety community to identify failure modes and compare mitigation strategies across frameworks in a reproducible manner. We release MASEval under an open-source licence to promote reproducible research and fair comparison in the multi-agent systems community.

\section*{Broader Impact Statement}

By providing standardised evaluation infrastructure for multi-agent systems, MASEval lowers the barrier to systematic benchmarking of both capability and safety properties. This dual applicability carries inherent tension. On one hand, MASEval can accelerate the development of more effective agentic systems. These systems inherit the risks associated with autonomous AI action, including compounding errors across agents and reduced human oversight in multi-agent coordination. On the other hand, the absence of rigorous, framework-agnostic evaluation tools makes it harder for the safety community to identify failure modes and compare mitigation strategies across implementations. We believe that standardised evaluation is a prerequisite for responsible deployment: systems that cannot be systematically measured cannot be systematically improved.

MASEval currently integrates four frameworks and seven benchmarks, and broader coverage depends on community contributions. For MASEval to serve as shared evaluation infrastructure at scale, this coverage must grow substantially. Our architecture has been tested with up to five agents. Tracing overhead grows linearly with the number of agents, but trace collection is negligible compared to the cost of running the agents themselves, which dominate wall-clock time and API spend. Scaling behaviour to larger multi-agent systems remains unexplored, and understanding these limits is important as real-world deployments move toward more complex agent topologies.

On the adapter side, MASEval's minimal interface (two methods) is designed as a lower bound. In practice, frameworks differ substantially in their abstractions for memory management, asynchronous execution, and tool calling. Some require thicker adapters that bridge these differences to MASEval's lifecycle model. The current interface has proven sufficient for four frameworks spanning diverse design philosophies, but novel architectures may require extending the adapter contract. How well this minimal contract accommodates future frameworks will determine whether MASEval can remain a unifying layer as the ecosystem evolves.

By design, MASEval prioritises flexibility over convenience. Its lightweight abstractions require more manual work than end-to-end evaluation libraries, a deliberate trade-off to support a wide range of benchmark designs. This entry barrier may slow adoption outside research settings. MASEval's primary audience is the research community. We expect that benchmark producers and consumers will benefit most from shared infrastructure that makes safety-relevant findings reproducible and comparable across the fragmented multi-agent landscape.

\section*{Acknowledgments}
This work was supported by the NAVER corporation. AR thanks the International Max Planck Research School for Intelligent Systems (IMPRS-IS) and Tübingen AI center for support.

\section*{Contribution Statement}


\noindent \textbf{C. Emde} conceived and implemented the MASEval library (core abstractions, framework interfaces, and all benchmark implementations unless noted otherwise), built the DevOps infrastructure, and wrote the documentation. He drafted the experimental settings, ran all experiments, analyzed results, prepared visualizations, reviewed related work, and drafted the paper.

\noindent \textbf{M. Gubri} provided daily supervision, helped refine the experimental settings, consolidated the paper's objectives, offered technical support, contributed to the initial and final draft.

\noindent \textbf{A. Goel} contributed the \textsc{ConVerse} benchmark and provided feedback throughout.

\noindent \textbf{A. Rubinstein} contributed the task queue for informative subsets for efficient agent evaluation, the DISCO example, and the MMLU and MMLU-Pro benchmarks.

\noindent \textbf{A. Heakl} contributed the ColBench benchmark.

\noindent \textbf{M. Gubri}, \textbf{S. J. Oh}, and \textbf{S. Yun} provided weekly supervision and supported the project through organizational and funding contributions.

\bibliography{references}

\clearpage
\appendix
\section*{\Large Appendix}

\section{Benchmarks and Frameworks}
\label{sec:supported}

Tables~\ref{tab:supported_benchmarks} and~\ref{tab:supported_frameworks} report the list of supported benchmarks and agentic frameworks.

\input{assets/tables/supported_benchmarks}
\input{assets/tables/supported_frameworks}

\section{Key Features}
\label{sec:features}

\begin{itemize}
    \item \textbf{Agent framework-agnostic.} Permissive abstract base class with documentation for custom adapters, plus pre-built adapters for smolagents, LangGraph, and LlamaIndex. Adding a new framework requires implementing only two methods: \code{\_run\_agent()} and \code{get\_messages()}.

    \item \textbf{Multi-agent native.} Built for multi-agent systems from the ground up with automatic collection of per-agent message histories, maintaining independent conversation contexts that respect partial observability. Each agent sees only its own messages, not other agents' internal states.

    \item \textbf{Comprehensive tracing.} Context-specific trace collection via \code{TraceableMixin}: agents log steps and message histories; models log input/output pairs, token usage, and latency; tools log invocations with inputs, outputs, and status; simulators log generation attempts and retries. Each component type captures what matters for its role.

    \item \textbf{Benchmark lifecycle management.} The \code{Benchmark} base class orchestrates a structured execution flow with overridable hooks: \code{setup\_environment()}, \code{setup\_agents()}, \code{setup\_user()}, \code{setup\_evaluators()}, \code{run\_agents()}, and \code{evaluate()}. Users override specific hooks while inheriting orchestration logic.

    \item \textbf{Callback hook system.} Abstract callback base classes (\code{BenchmarkCallback}, \code{EnvironmentCallback}, \code{AgentCallback}) provide lifecycle hooks at benchmark, environment, and agent levels; built-in callbacks include progress bars (tqdm, Rich) and result logging.

    \item \textbf{Environment abstraction.} Abstract \code{Environment} base class for custom task environments; users implement \code{setup\_state()} and \code{create\_tools()} to define environment initialisation and available tools.

    \item \textbf{Standardised two-stage evaluation.} Abstract \code{Evaluator} base class with \code{filter\_traces()} to extract relevant data (e.g., specific tool calls) before \code{\_\_call\_\_()} computes metrics; this separation enables reusable evaluation logic across different trace sources.

    \item \textbf{Custom execution loop.} The \code{run\_agents()} method can be overridden to implement custom turn-taking strategies: user-initiated (user speaks first), model-initiated (agent speaks first), alternating, or fully custom interaction patterns. Default implementation supports configurable multi-turn agent-user exchanges.

    \item \textbf{Multi-turn user simulation.} Abstract \code{User} base class for custom user simulation logic, plus pre-built \code{UserLLMSimulator} for LLM-based response generation. Configurable maximum turns, stop tokens for early termination, and automatic conversation history tracking. Supports both message-based turn-taking (standard chatbot interaction) and tool-based interaction for frameworks that model user queries as tool calls (e.g., \code{ask\_user}).

    \item \textbf{Parallel execution.} The \code{num\_workers} parameter enables concurrent task execution via thread pool. All library components are thread-safe: per-thread component registries prevent cross-contamination, locks serialise callback invocations and report aggregation, and trace collection is designed for concurrent access.

    \item \textbf{Structured error attribution.} Exception hierarchy distinguishes \code{AgentError} (agent violated contract, i.e., counts against score) from \code{EnvironmentError} and \code{UserError} (infrastructure failures, i.e., excluded from scoring). This prevents penalising agents for benchmark bugs. Developers can raise \code{AgentError} with an optional \code{suggestion} field; custom agent implementations can catch these errors and use the suggestion to retry with corrected inputs.


    \item \textbf{Component registry.} Thread-safe registration of agents, models, tools, and simulators via \code{ComponentRegistry}. Components returned from setup methods are automatically registered; additional components can be manually registered. Enables systematic trace and config collection across all registered components.

    \item \textbf{Unified model interface.} Abstract base class \code{ModelAdapter} for custom LLM providers, plus pre-built adapters for OpenAI, Google, Anthropic, and HuggingFace with automatic token tracking and tool-calling support.

    \item \textbf{LLM simulators.} Abstract simulator base classes with pre-built \code{ToolLLMSimulator} and \code{UserLLMSimulator} that generate realistic tool responses and user turns when real APIs are unavailable, enabling offline development and testing.

    \item \textbf{Configuration snapshotting.} The \code{ConfigurableMixin} can be added to any user-defined class to participate in configuration capture for reproducibility; built-in collection includes Git state, system information, and package versions.

    \item \textbf{Pluggable logging backends.} Abstract callback base class enables routing results and traces to any destination; JSON file logging is pre-built, with documentation for WandB and Langfuse integration.

    \item \textbf{Debugging tools.} Configurable error handling lets benchmarks continue on failures (\code{fail\_on\_task\_error=False}) for batch runs or fail fast (\code{fail\_on\_task\_error=True}) for interactive debugging. Failed tasks are tracked automatically and can be retried selectively.

    \item \textbf{Adaptive testing.} Abstract \code{TaskQueue} base class enables custom task selection strategies such as Item-Response Theory-based testing or DISCO \citep{rubinstein2026disco} to reduce evaluation costs; priority-based and sequential queues are pre-implemented.

    \item \textbf{Robust task execution with timeout handling and repetition.} The \code{n\_task\_repeats} parameter runs each task multiple times for statistical robustness. Reports include \code{repeat\_idx} for aggregation across runs. Per-task deadlines via \code{TaskContext} with cooperative checkpoint-based timeout (not forced thread termination). Configurable timeout actions: skip, retry, or extend.

    \item \textbf{Structured task protocol.} Each \code{Task} carries metadata including timeout configuration, retry policies, priority levels, and custom tags for fine-grained control over execution behaviour.
\end{itemize}

\end{document}

%% file: custom_commands.tex
\usepackage{soul}
\usepackage{xcolor}

\definecolor{todocolor}{RGB}{255, 255, 150}  

\usepackage{tcolorbox}
\tcbuselibrary{listings, skins, breakable}

\definecolor{inlinecodebg}{RGB}{232, 232, 232}
\definecolor{inlinecodefg}{RGB}{68, 38, 120}
\newtcbox{\code}{
  on line,
  colback=inlinecodebg,
  colframe=inlinecodebg,
  boxrule=0pt,
  arc=2.5pt,
  boxsep=0pt,
  left=2pt, right=2pt, top=1.5pt, bottom=1pt,
  fontupper=\ttfamily\color{inlinecodefg}
}

\definecolor{codebg}{RGB}{248, 248, 248}
\definecolor{codeborder}{RGB}{220, 220, 220}
\definecolor{codekeyword}{RGB}{0, 119, 170}
\definecolor{codestring}{RGB}{170, 85, 0}
\definecolor{codecomment}{RGB}{100, 100, 100}

\usepackage{pifont}
\definecolor{tblgreen}{RGB}{34, 139, 34}
\definecolor{tblteal}{RGB}{0, 128, 128}
\definecolor{tblyellow}{RGB}{204, 153, 0}
\definecolor{tblred}{RGB}{178, 34, 34}
\definecolor{tblgray}{RGB}{128, 128, 128}
\newcommand{\cmark}{{\color{tblgreen}\ding{51}}}  
\newcommand{\xmark}{{\color{tblred}\ding{55}}}    
\newcommand{\pmark}{{\color{tblyellow}$\circ$}}   

\usepackage{enumitem}
\setlist{
  topsep=3pt,       
  itemsep=2pt,      
  parsep=1pt,       
  leftmargin=1.5em  
}
\definecolor{defblue}{RGB}{0, 80, 160}

\newtcblisting{codebox}{
    enhanced,
    listing only,
    sharp corners,
    boxrule=0.4pt,
    colback=codebg,
    colframe=codeborder,
    width=\linewidth,
    left=1pt,
    right=1pt,
    top=0pt,
    bottom=0pt,
    boxsep=1pt,
    listing options={
        language=Python,
        basicstyle=\ttfamily\scriptsize,
        breaklines=true,
        showstringspaces=false,
        keywordstyle=\color{codekeyword}\bfseries,
        stringstyle=\color{codestring},
        commentstyle=\color{codecomment}\itshape,
        columns=fixed,
        keepspaces=true,
        frame=none,
        aboveskip=0pt,
        belowskip=0pt
    }
}

%% file: assets/figures/motivating_figure.tex
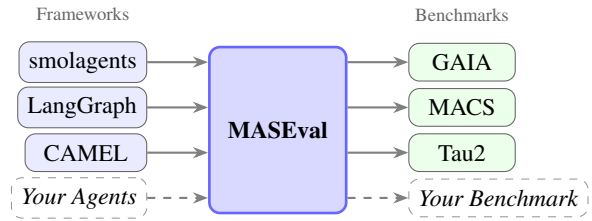
\begin{figure}[t]
\vspace{-15pt}
\centering
\begin{tikzpicture}[
    node distance=0.35cm,
    framework/.style={rectangle, rounded corners, draw=black!60, fill=blue!8, minimum width=1.6cm, minimum height=0.5cm, font=\footnotesize},
    benchmark/.style={rectangle, rounded corners, draw=black!60, fill=green!8, minimum width=1.4cm, minimum height=0.5cm, font=\footnotesize},
    custom/.style={rectangle, rounded corners, draw=black!40, dashed, fill=white, minimum width=1.4cm, minimum height=0.5cm, font=\footnotesize\itshape},
    maseval/.style={rectangle, rounded corners, draw=blue!60, fill=blue!15, line width=1pt, minimum width=1.8cm, minimum height=2.2cm, font=\footnotesize\bfseries},
    install/.style={rectangle, rounded corners, draw=black!50, fill=gray!10, minimum width=2.6cm, minimum height=0.5cm, font=\footnotesize\ttfamily},
    arrow/.style={-{Stealth[length=2mm]}, thick, draw=black!50}
]


\node[maseval] (maseval) {MASEval};

\node[framework, left=0.8cm of maseval, yshift=0.9cm] (smol) {smolagents};
\node[framework, left=0.8cm of maseval, yshift=0.3cm] (lang) {LangGraph};
\node[framework, left=0.8cm of maseval, yshift=-0.3cm] (auto) {CAMEL};
\node[custom, left=0.8cm of maseval, yshift=-0.9cm] (youragent) {Your Agents};

\node[benchmark, right=0.8cm of maseval, yshift=0.9cm] (gaia) {GAIA};
\node[benchmark, right=0.8cm of maseval, yshift=0.3cm] (macs) {MACS};
\node[benchmark, right=0.8cm of maseval, yshift=-0.3cm] (tau) {Tau2};
\node[custom, right=0.8cm of maseval, yshift=-0.9cm] (yourbench) {Your Benchmark};

\draw[arrow] (smol.east) -- (maseval.west |- smol.east);
\draw[arrow] (lang.east) -- (maseval.west |- lang.east);
\draw[arrow] (auto.east) -- (maseval.west |- auto.east);
\draw[arrow, dashed] (youragent.east) -- (maseval.west |- youragent.east);

\draw[arrow] (maseval.east |- gaia.west) -- (gaia.west);
\draw[arrow] (maseval.east |- macs.west) -- (macs.west);
\draw[arrow] (maseval.east |- tau.west) -- (tau.west);
\draw[arrow, dashed] (maseval.east |- yourbench.west) -- (yourbench.west);

\node[font=\scriptsize, above=0.15cm of smol, text=black!60] {Frameworks};
\node[font=\scriptsize, above=0.15cm of gaia, text=black!60] {Benchmarks};

\end{tikzpicture}
\vspace{-2pt}
\caption{MASEval provides a unified evaluation layer that enables framework-agnostic, system-level comparison across any agent framework and benchmark.}
\vspace{-10pt}
\label{fig:motivation}
\end{figure}

%% file: assets/tables/library_comparison.tex
\begin{table*}[t] 
\centering
\resizebox{\textwidth}{!}{%
\begin{tabular}{lcccccccc}
\toprule
\textbf{Library} & \textbf{Multi-Agent} & \textbf{System Eval} & \textbf{Agent-Agnostic} & \textbf{Benchmarks} & \textbf{Flexible Interaction} & \textbf{BYO} & \textbf{Trace-First} & \textbf{Mature} \\
\midrule
\textbf{MASEval (Ours)} & \cmark & \cmark & \cmark & \cmark & \cmark & \cmark & \cmark & \cmark \\
AnyAgent \citep{any_agent} & \pmark & \cmark & \cmark & \xmark & \pmark & \cmark & \pmark & \cmark \\
MLflow GenAI \citep{zaharia2018mlflow} & \pmark & \pmark & \cmark & \xmark & \pmark & \cmark & \cmark & \cmark \\
HAL Harness \citep{kapoor2026hal} & \pmark & \cmark & \cmark & \cmark & \pmark & \pmark & \pmark & \pmark \\
Inspect-AI \citep{inspectai2024} & \pmark & \cmark & \pmark & \cmark & \pmark & \pmark & \pmark & \cmark \\
OpenCompass \citep{2023opencompass} & \xmark & \pmark & \xmark & \cmark & \pmark & \pmark & \pmark & \cmark \\
AgentGym \citep{xi2024agentgym} & \xmark & \xmark & \xmark & \cmark & \pmark & \cmark & \pmark & \pmark \\
Arize Phoenix \citep{phoenix2022} & \pmark & \xmark & \pmark & \xmark & \xmark & \pmark & \cmark & \cmark \\
TruLens \citep{trulens2020} & \pmark & \xmark & \pmark & \xmark & \xmark & \pmark & \cmark & \cmark \\
MARBLE \citep{zhu2025multiagentbench} & \cmark & \xmark & \xmark & \cmark & \xmark & \xmark & \pmark & \pmark \\
DeepEval \citep{deepeval2026} & \pmark & \xmark & \pmark & \xmark & \pmark & \pmark & \pmark & \cmark \\
MCPEval \citep{mcpeval2025} & \xmark & \xmark & \xmark & \cmark & \xmark & \pmark & \pmark & \pmark \\
\bottomrule
\end{tabular}%
}
\vspace{-4pt} 
\caption{Comparison with related libraries. \cmark~= built-in support or explicitly designed to be extensible, \pmark~= partial support, \xmark~= not supported/impractical. \textbf{Multi-Agent}: native orchestration with per-agent tracing and independent message histories. \textbf{System-Level}: compare framework implementations, not just LLMs. \textbf{Agent-Agnostic}: evaluate any framework via thin adapters without code recreation. \textbf{Benchmarks}: ships complete, ready-to-run benchmarks. \textbf{Flexible Interaction}: Flexible Agent-Environment-User Interaction. First-class user simulation with personas and tool access. \textbf{BYO}: bring your own logging, agents, environments, and tools. Open-source, works offline, no mandatory cloud services. \textbf{Trace-First}: evaluate intermediate steps across environment and agents via first-class traces, not post-hoc fixes. \textbf{Mature}: published on PyPI, CI/CD, good test coverage, active maintenance.}
\label{tab:comparison}
\vspace{-10pt}
\end{table*}

%% file: assets/figures/byo_philosophy.tex
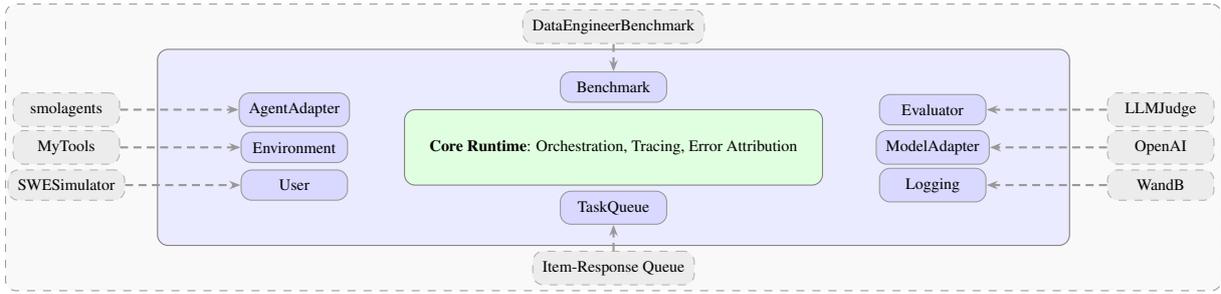
\begin{figure*}[t]
\centering
\begin{tikzpicture}[
    node distance=0.2cm,
    layer/.style={rectangle, draw=black!50, rounded corners, font=\scriptsize, align=center},
    comp/.style={rectangle, draw=black!40, rounded corners, minimum width=1.4cm, minimum height=0.4cm, font=\tiny, align=center},
    arrow/.style={-{Stealth[length=1.5mm]}, thick, draw=black!40, densely dashed}
]

\node[layer, draw=black!40, dashed, fill=gray!5, minimum width=16cm, minimum height=3.8cm] (user) {};

\node[layer, fill=blue!8, minimum width=12cm, minimum height=2.6cm] (maseval) {};

\node[layer, fill=green!12, minimum width=5.5cm, minimum height=1.0cm] (core) {};
\node[font=\tiny, align=center] at (core.center) {\textbf{Core Runtime}: Orchestration, Tracing, Error Attribution};

\node[comp, fill=blue!15] at ([yshift=0.8cm]core.center) (bc7) {Benchmark};
\node[comp, fill=gray!15, dashed, draw=black!30, minimum width=2.2cm] at ([yshift=1.6cm]core.center) (u7) {DataEngineerBenchmark};

\node[comp, fill=blue!15] at ([xshift=-4.2cm, yshift=0.5cm]core.center) (bc1) {AgentAdapter};
\node[comp, fill=blue!15] at ([xshift=-4.2cm, yshift=0cm]core.center) (bc2) {Environment};
\node[comp, fill=blue!15] at ([xshift=-4.2cm, yshift=-0.5cm]core.center) (bc3) {User};
\node[comp, fill=blue!15] at ([xshift=4.2cm, yshift=0.5cm]core.center) (bc4) {Evaluator};
\node[comp, fill=blue!15] at ([xshift=4.2cm, yshift=0cm]core.center) (bc5) {ModelAdapter};
\node[comp, fill=blue!15] at ([xshift=4.2cm, yshift=-0.5cm]core.center) (bc6) {Logging};

\node[comp, fill=blue!15] at ([yshift=-0.8cm]core.center) (bc8) {TaskQueue};
\node[comp, fill=gray!15, dashed, draw=black!30, minimum width=2cm] at ([yshift=-1.6cm]core.center) (u8) {Item-Response Queue};

\node[comp, fill=gray!15, dashed, draw=black!30] at ([xshift=-7.2cm, yshift=0.5cm]core.center) (u1) {smolagents};
\node[comp, fill=gray!15, dashed, draw=black!30] at ([xshift=-7.2cm, yshift=0cm]core.center) (u2) {MyTools};
\node[comp, fill=gray!15, dashed, draw=black!30] at ([xshift=-7.2cm, yshift=-0.5cm]core.center) (u3) {SWESimulator};
\node[comp, fill=gray!15, dashed, draw=black!30] at ([xshift=7.2cm, yshift=0.5cm]core.center) (u4) {LLMJudge};
\node[comp, fill=gray!15, dashed, draw=black!30] at ([xshift=7.2cm, yshift=0cm]core.center) (u5) {OpenAI};
\node[comp, fill=gray!15, dashed, draw=black!30] at ([xshift=7.2cm, yshift=-0.5cm]core.center) (u6) {WandB};

\draw[arrow] (u1.east) -- (bc1.west);
\draw[arrow] (u2.east) -- (bc2.west);
\draw[arrow] (u3.east) -- (bc3.west);
\draw[arrow] (u4.west) -- (bc4.east);
\draw[arrow] (u5.west) -- (bc5.east);
\draw[arrow] (u6.west) -- (bc6.east);
\draw[arrow] (u7.south) -- (bc7.north);
\draw[arrow] (u8.north) -- (bc8.south);

\end{tikzpicture}
\caption{MASEval adopts a ``Bring Your Own'' (BYO) philosophy. Users implement \colorbox{gray!15}{custom components} by extending MASEval's \colorbox{blue!15}{abstract base classes}; the \colorbox{green!15}{core runtime} orchestrates execution and collects traces. This enables maximum flexibility while minimizing boilerplate code.}
\label{fig:byo-philosophy}
\end{figure*}

%% file: assets/figures/benchmark_lifecycle.tex
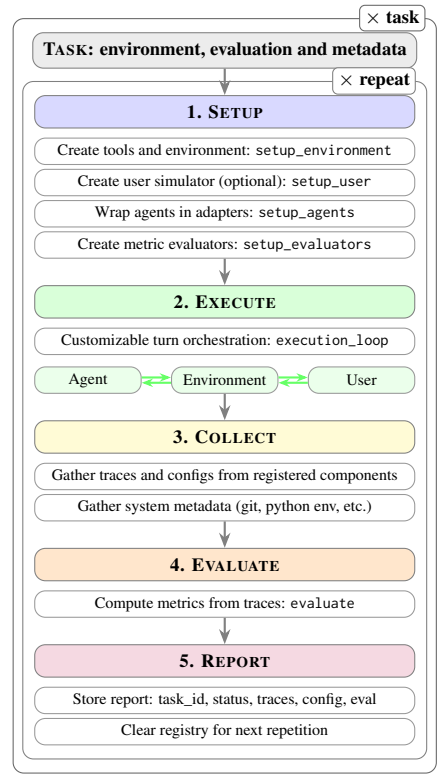
\begin{figure}[t]
\vspace{-10pt}
\centering
\begin{tikzpicture}[
    node distance=0.12cm,
    phase/.style={rectangle, draw=black!50, fill=#1, rounded corners, minimum width=5.0cm, minimum height=0.4cm, font=\scriptsize\bfseries, align=center},
    hook/.style={rectangle, draw=black!40, fill=white, rounded corners, minimum width=5.0cm, minimum height=0.35cm, font=\tiny, align=center, inner sep=2pt},
    arrow/.style={-{Stealth[length=2mm]}, thick, draw=black!50},
    boxtitle/.style={rectangle, draw=black!50, fill=white, rounded corners=2pt, font=\scriptsize\bfseries, inner sep=2pt}
]

\node[phase=gray!15] (task) {\textsc{Task}: environment, evaluation and metadata};

\node[phase=blue!15, below=0.35cm of task] (p1) {\textsc{1. Setup}};
\node[hook, below=0.12cm of p1] (h1) {Create tools and environment: \texttt{setup\_environment}};
\node[hook, below=0.05cm of h1] (h2) {Create user simulator (optional): \texttt{setup\_user}};
\node[hook, below=0.05cm of h2] (h3) {Wrap agents in adapters: \texttt{setup\_agents}};
\node[hook, below=0.05cm of h3] (h4) {Create metric evaluators: \texttt{setup\_evaluators}};

\node[phase=green!15, below=0.35cm of h4] (p2) {\textsc{2. Execute}};
\node[hook, below=0.12cm of p2] (e1) {Customizable turn orchestration: \texttt{execution\_loop}};
\node[hook, fill=green!8, below=0.15cm of e1, minimum width=1.4cm, xshift=-1.8cm] (eagent) {Agent};
\node[hook, fill=green!8, below=0.15cm of e1, minimum width=1.4cm] (eenv) {Environment};
\node[hook, fill=green!8, below=0.15cm of e1, minimum width=1.4cm, xshift=1.8cm] (euser) {User};
\draw[-{Stealth[length=1.5mm]}, thick, draw=green!60] ([yshift=1pt]eagent.east) -- ([yshift=1pt]eenv.west);
\draw[-{Stealth[length=1.5mm]}, thick, draw=green!60] ([yshift=-1pt]eenv.west) -- ([yshift=-1pt]eagent.east);
\draw[-{Stealth[length=1.5mm]}, thick, draw=green!60] ([yshift=1pt]eenv.east) -- ([yshift=1pt]euser.west);
\draw[-{Stealth[length=1.5mm]}, thick, draw=green!60] ([yshift=-1pt]euser.west) -- ([yshift=-1pt]eenv.east);

\node[phase=yellow!20, below=0.35cm of eenv] (p3) {\textsc{3. Collect}};
\node[hook, below=0.12cm of p3] (c1) {Gather traces and configs from registered components};
\node[hook, below=0.05cm of c1] (c2) {Gather system metadata (git, python env, etc.)};

\node[phase=orange!20, below=0.35cm of c2] (p4) {\textsc{4. Evaluate}};
\node[hook, below=0.12cm of p4] (v1) {Compute metrics from traces: \texttt{evaluate}};

\node[phase=purple!15, below=0.35cm of v1] (p5) {\textsc{5. Report}};
\node[hook, below=0.12cm of p5] (r1) {Store report: task\_id, status, traces, config, eval};
\node[hook, below=0.05cm of r1] (r2) {Clear registry for next repetition};

\node[rectangle, draw=black!50, rounded corners, inner xsep=0.15cm, inner ysep=0.18cm,
      fit=(p1)(r2)] (repeatloop) {};
\node[boxtitle, anchor=east] at ([xshift=-0.15cm]repeatloop.north east) {$\times$ repeat};

\node[rectangle, draw=black!60, rounded corners, inner xsep=0.12cm, inner ysep=0.18cm,
      fit=(task)(repeatloop)] (taskloop) {};
\node[boxtitle, draw=black!60, anchor=east] at ([xshift=-0.15cm]taskloop.north east) {$\times$ task};

\draw[arrow] (task.south) -- (p1.north); 
\draw[arrow] (h4.south) -- (p2.north);
\draw[arrow] (eenv.south) -- (p3.north);
\draw[arrow] (c2.south) -- (p4.north);
\draw[arrow] (v1.south) -- (p5.north);

\end{tikzpicture} 
\vspace{-5pt}
\caption{Benchmark task lifecycle with flexible execution. The outer loop iterates over tasks; the inner loop handles repetitions. The Execute phase shows agent-user interaction as a flexible bidirectional loop.}
\label{fig:benchmark-lifecycle}
\vspace{-10pt}
\end{figure}

%% file: assets/tables/framework_performance.tex
\begin{table*}[ht]
\centering
\resizebox{\textwidth}{!}{%
\begin{tabular}{@{}ll rrrrrr@{}}
\toprule
 & & \multicolumn{2}{c}{\textbf{MACS} $\uparrow$} & \multicolumn{2}{c}{\textbf{\textsc{ConVerse}} $\uparrow$} & \multicolumn{2}{c@{}}{\textbf{MultiAgentBench} $\uparrow$} \\
\cmidrule(lr){3-4} \cmidrule(lr){5-6} \cmidrule(l){7-8}
\textbf{Framework} & \textbf{Model} & Travel & Mortgage & Travel Planning & Real Estate & Research & Bargaining \\
\midrule
\multirow{3}{*}{smolagents} & Gemini-3.0-Flash &  84.0 & \textbf{94.4} &  84.8 &  82.4 &  99.0 &  89.0 \\
                            & GPT-5-mini       &  59.8 &  85.8 &  90.1 &  84.2 &  98.1 &  91.7 \\
                            & Haiku 4.5        & \textbf{90.4} &  85.6 &  86.2 &  86.0 & \textbf{100.0} &  90.2 \\
\midrule
\multirow{3}{*}{LangGraph}  & Gemini-3.0-Flash &  85.8 &  89.4 &  60.0 &  72.2 &  98.0 &  82.7 \\
                            & GPT-5-mini       &  60.8 &  73.7 &  73.3 &  77.0 &  95.5 & \textbf{93.6} \\
                            & Haiku 4.5        &  68.3 &  81.2 & \textbf{95.8} &  96.7 &  92.4 &  87.1 \\
\midrule
\multirow{3}{*}{LlamaIndex} & Gemini-3.0-Flash &  74.7 &  93.2 &  93.8 & \textbf{100.0} &  96.0 &  75.6 \\
                            & GPT-5-mini       &  71.0 &  76.7 &  85.4 &  75.3 &  92.0 &  90.8 \\
                            & Haiku 4.5        &  59.5 &  76.7 &  94.7 &  97.4 &  95.8 &  87.1 \\
\midrule
\midrule
\multirow{2}{*}{\textit{Mean Range}} & \textit{Cross-Model}     & 23.6 & 13.7 & 16.8 & 17.6 &  3.8 &  9.6 \\
                               & \textit{Cross-Framework} & 17.7 &  8.7 & 20.1 & 16.0 &  5.6 &  6.4 \\
\midrule
\multirow{2}{*}{\textit{Mean SD}}  & \textit{Cross-Model}        & 12.3 &  7.5 &  8.7 &  9.5 &  2.0 &  4.9 \\
                               & \textit{Cross-Framework}    &  9.4 &  4.5 & 10.5 &  8.4 &  2.8 &  3.3 \\
\bottomrule
\end{tabular}%
}
\vspace{-4pt}
\caption{Performance across frameworks, models, benchmarks, and domains. ConVerse columns shows $(1- \text{ASR})$ on security split. MultiAgentBench Research reports Completion Rate and Bargaining reports Task Score. Bold marks the best result per task column. \textit{Cross-Model} statistics measure variability across models within each framework, averaged over frameworks. \textit{Cross-Framework} statistics measure the reverse.}
\label{tab:results}
\vspace{-10pt}
\end{table*}

%% file: assets/tables/loc_comparison.tex
\begin{table}[t]
\centering
\resizebox{\columnwidth}{!}{%
\begin{tabular}{@{}llrr rr@{}}
\toprule
 & & & & \multicolumn{2}{c}{\textbf{Change}} \\
\cmidrule(l){5-6}
\textbf{Benchmark} & \textbf{Component} & \textbf{Original} & \textbf{MASEval} & \textbf{LoC} & \textbf{\%} \\
\midrule
\multirow{3}{*}{$\tau^2$-Bench} & Definition & 6,822 & 3,450 & $-$3,372 & $-$49.4 \\
 & Interface & 1,982 & 343 & $-$1,639 & $-$82.7 \\
 & Total & 8,804 & 3,793 & $-$5,011 & $-$56.9 \\
\midrule
\multirow{3}{*}{\textsc{ConVerse}} & Definition & 1,320 & 1,283 & $-$37 & $-$2.8 \\
 & Interface & 778 & 71 & $-$707 & $-$90.9 \\
 & Total & 2,098 & 1,354 & $-$744 & $-$35.5 \\
\bottomrule
\end{tabular}%
}
\caption{Lines of code comparison for benchmark evaluation logic. \emph{Definition} covers task specifications, environments, and evaluators. \emph{Interface} covers the CLI and entry points that wire components into a runnable benchmark. MASEval replaces framework- and benchmark-specific orchestration with shared abstractions.}
\label{tab:loc}
\vspace{-10pt}
\end{table}

%% file: assets/tables/supported_benchmarks.tex
\begin{table}[h]
\centering
\resizebox{\columnwidth}{!}{%
\begin{tabular}{lll}
\toprule
\textbf{Benchmark} & \textbf{Type} & \textbf{Domain} \\
\midrule
GAIA-2 \citep{mialon2024gaia} & SA & Capability \\
$\tau^2$-bench \citep{barres2025tau2} & SA & Capability \\
\textsc{MMLU} \citep{hendryckstest2021} & SA & Capability \\
MACS \citep{shu2024macs} & MA & Collaboration \\
MultiAgentBench \citep{zhu2025multiagentbench} & MA & Coordination \& Competition \\
\textsc{ConVerse} \citep{gomaa2025converse} & MA & Safety \& Security \\
\textsc{ColBench} \citep{zhou2025sweetrltrainingmultiturnllm} & MA & Capability \\
\bottomrule
\end{tabular}%
}
\caption{Currently supported benchmarks. SA$=$Single-Agent, MA$=$Multi-agent.}
\label{tab:supported_benchmarks}
\end{table}

%% file: assets/tables/supported_frameworks.tex
\begin{table}[h]
\centering
\resizebox{\columnwidth}{!}{%
\begin{tabular}{ll}
\toprule
\textbf{Framework} & \textbf{Architecture} \\
\midrule
smolagents \citep{smolagents} & Code-based tool calling \\
LangGraph \citep{langgraph} & Stateful graph execution \\
LlamaIndex \citep{llamaindex2022} & Async-first workflows \\
CAMEL \citep{li2023camel} & Role-playing multi-agent \\
\bottomrule
\end{tabular}%
}
\caption{Currently supported agentic frameworks. Each framework is integrated via a thin adapter that exposes a unified interface for execution, message history, tracing, and configuration capture.}
\label{tab:supported_frameworks}
\end{table}